\documentclass[11pt]{article} 
\usepackage{rldm,palatino}
\usepackage{paper_header}

\title{Adaptive Tree Backup Algorithms for\\Temporal-Difference Reinforcement Learning}

\author{
Brett Daley\thanks{Equal contribution.}\\
Khoury College of Computer Sciences\\
Northeastern University\\
Boston, MA 02115\\
\texttt{daley.br@northeastern.edu} \\
\And
Isaac Chan\footnotemark[1]\\
Khoury College of Computer Sciences\\
Northeastern University\\
Boston, MA 02115\\
\texttt{chan.is@northeastern.edu}\\
}

\begin{document}

\maketitle

\begin{abstract}
    Q($\sigma$) is a recently proposed temporal-difference learning method that
    interpolates between learning from expected backups and sampled backups.
    It has been shown that intermediate values for the interpolation parameter $\sigma \in [0,1]$ perform better in practice, and therefore it is commonly believed that $\sigma$ functions as a bias-variance trade-off parameter to achieve these improvements.
    In our work, we disprove this notion, showing that the choice of $\sigma=0$ minimizes variance without increasing bias.
    This indicates that $\sigma$ must have some other effect on learning that is not fully understood.
    As an alternative, we hypothesize the existence of a new trade-off:
    larger $\sigma$-values help overcome poor initializations of the value function, at the expense of higher statistical variance.
    To automatically balance these considerations, we propose Adaptive Tree Backup (ATB) methods, whose weighted backups evolve as the agent gains experience.
    Our experiments demonstrate that adaptive strategies can be more effective than relying on fixed or time-annealed $\sigma$-values.
\end{abstract}

\keywords{
Reinforcement Learning, Temporal-Difference Learning,\\Adaptive Methods, Tree Backup, Expected Sarsa
}


\startmain 

\tightsection{Introduction}

Unifying temporal-difference (TD), dynamic programming (DP), and Monte Carlo (MC) methods has helped illuminate trade-offs in reinforcement learning and has lead to the development of better algorithms.
Much work has focused on the \textit{length} of backups as a unifying axis between TD and MC methods, in which these two classes can be understood as opposite extremes of the TD($\lambda$) algorithm \citep{sutton1988learning, sutton1998reinforcement}.
In this view, the decay parameter $\lambda \in [0,1]$ serves not only as a conceptual link between these distinct methods, but also as a mechanism for managing the bias-variance trade-off \citep{kearns2000bias}, where intermediate values of $\lambda$ often deliver the best empirical performance \citep{sutton1998reinforcement}.

Recent research has begun to explore an orthogonal unification between TD and DP methods by varying the \textit{width} of backups.
This concept was formally introduced by \citet{sutton2018reinforcement} in the Q($\sigma$) algorithm;
the method interpolates linearly, via a parameter $\sigma \in [0,1]$, between Expected Sarsa ($\sigma=0$) \citep{sutton1998reinforcement} and Sarsa ($\sigma=1$) \citep{rummery1994line}.
As such, Q($\sigma$) can be seen as a spectrum of algorithms whose backup widths lie somewhere between those of a sample backup and a full (expected) backup.
\citet{de2018multi} conducted further theoretical and empirical analysis of Q($\sigma$), arriving at two major hypotheses:
$\sigma$ is a bias-variance trade-off parameter, and dynamically adapting $\sigma$ can lead to faster convergence.
These hypotheses appeared to be corroborated by their experiments, where intermediate $\sigma$-values performed well, and exponentially decaying $\sigma$ during training performed even better.

In our work, we re-examine these two hypotheses regarding Q($\sigma$), and ultimately show that they are incorrect.
In particular, we prove that the choice of $\sigma=0$ minimizes the variance of Q($\sigma$) without increasing the bias.
It follows that the choice of $\sigma$ does not constitute a bias-variance trade-off, since both bias and variance can be simultaneously optimized without competition.
Furthermore, we prove that the contraction rate and fixed point of the Q($\sigma$) operator are independent of $\sigma$, suggesting again that the choice of $\sigma=0$ is preferable for its variance-reduction effect.
These findings are intriguing given the empirical results obtained by \citet{de2018multi}, which did not demonstrate a clear superiority of $\sigma=0$.
This apparent discrepancy between theory and practice suggests that other factors are responsible for the observed benefits of Q($\sigma$), and that our understanding of the algorithm must be revised accordingly.

We therefore advance a new hypothesis to explain the effectiveness of intermediate $\sigma$-values:
conducting partial backups ($\sigma \neq 0$) helps escape poor initializations of the value function $Q$ by reducing the chance that previously unvisited state-action pairs are considered during the backup.
The value estimates of these pairs could be arbitrarily incorrect, slowing initial learning;
however, as training progresses and the majority of state-action pairs have been visited, $\sigma=0$ starts to become more favorable for its variance reduction.
This explanation matches the intuition of the time-decayed schedule utilized by \citet{de2018multi}, in which $\sigma=1$ is desirable early in training and $\sigma=0$ is desirable later.

This new hypothesis suggests that backup width should be adjusted based on the number of visitations to each state-action pair, with more emphasis placed on frequently sampled pairs.
We therefore develop a new family of TD algorithms that we call Adaptive Tree Backup (ATB) methods, which generalizes Tree Backup \citep{precup2000eligibility} to arbitrary weightings of the value estimates.
We highlight two instances of ATB methods (\emph{Count-Based} and \emph{Policy-Based}) that we believe to be promising.
The proposed methods have the advantage of eliminating the hyperparameter $\sigma$, as they adapt automatically to the data experienced by the agent.
Finally, our experiments demonstrate that Policy-Based ATB outperforms the exponential-decay schedule for $\sigma$ introduced by \citet{de2018multi}.

\tightsection{Background}

We model the environment as a Markov Decision Process (MDP) described by the tuple $(\mathcal{S}, \mathcal{A}, P, R)$:
$\mathcal{S}$ is a finite set of environment states,
$\mathcal{A}$ is a finite set of actions,
$P$ is the transition function,
and $R$ is the reward function.
An agent interacts with the environment by selecting an action $a_t \in \mathcal{A}$ in state $s_t$ with probability $\pi(a_t|s_t)$, receiving a reward $r_t \coloneqq R(s_t,a_t)$ and changing the state to $s_{t+1}$ with probability $P(s_{t+1}|s_t,a_t)$.
Given the agent's policy $\pi$, the objective is to estimate the expected discounted return $Q^\pi(s_t,a_t) \coloneqq \E_\pi[r_t + \gamma r_{t+1} + \gamma^2 r_{t+2} + \dots]$,
where $\gamma \in [0,1]$, since acting greedily with respect to $Q^\pi$ is known to produce a better policy \citep{sutton1998reinforcement}.

Q($\sigma$) \citep{sutton2018reinforcement} is a method that iteratively improves its estimated value function $Q$ according to
\begin{equation}
    \label{eq:qsigma_update}
    Q(s_t,a_t) \gets (1-\alpha_t) Q(s_t,a_t) + \alpha_t \left( r_t + \gamma \left( \sigma Q(s_{t+1}, a_{t+1}) + (1-\sigma) \sum_{a' \in \mathcal{A}} \pi(a'|s_{t+1}) Q(s_{t+1},a') \right) \right)
    ,
\end{equation}
where $\sigma \in [0,1]$ is a user-specified hyperparameter.
This hyperparameter interpolates between the exact on-policy expectation (Expected Sarsa, $\sigma=0$) and noisy sample estimates of it (Sarsa, $\sigma=1$), and has been hypothesized to serve as a bias-variance trade-off parameter that improves the contraction rate of the algorithm \citep{de2018multi}.
In the subsequent section, we analyze the theoretical properties of the Q($\sigma$) update to investigate the validity of this hypothesis.

\tightsection{Analysis of Q($\sigma$)}

We investigate the effects that $\sigma$ has on the Q($\sigma$) algorithm in terms of convergence, contraction rate, bias, and variance.
Our analysis dispels two misconceptions about Q($\sigma$):
that dynamically adjusting $\sigma$ can achieve a faster contraction rate, and that $\sigma$ functions as a bias-variance trade-off parameter.

\subsection{Convergence and Contraction Rate}

For notational conciseness when proving convergence, we represent the Q($\sigma$) update (\ref{eq:qsigma_update}) as an operator $T_\pi^\sigma \colon \mathbb{R}^n \to \mathbb{R}^n$, where $n \coloneqq |\mathcal{S} \times \mathcal{A}|$.
The Bellman operator $T_\pi$ for a policy $\pi$ is defined as the affine function $Q \mapsto R + \gamma P_\pi Q$,
where $\smash{(P_\pi Q)(s,a) \coloneqq \sum_{s',a'} P(s'|s,a) \pi(a'|s') Q(a'|s')}$.
The Bellman operator is a contraction mapping (when $\gamma < 1)$ that admits the unique fixed point $\smash{Q^\pi = \sum_{k=0}^\infty (\gamma P_\pi)^k R}$, the unique solution to $T_\pi Q = Q$ \citep{bellman1966dynamic}.

For policy evaluation, the operator $T_\pi^\sigma$ is considered convergent if and only if $\lim_{k \to \infty} (T_\pi^\sigma)^k Q = Q^\pi$ for every $Q \in \mathbb{R}^n$, where
$(T_\pi^\sigma)^k Q \coloneqq (T_\pi^\sigma)^{k-1} (T_\pi^\sigma Q)$.
Let $w \in \mathbb{R}^n$ be zero-mean noise that represents the randomness due to the environment and policy.
Since Q($\sigma$) interpolates linearly between Expected Sarsa and Sarsa, we can express its operator as
\begin{equation}
    \label{eq:qsigma_operator}
    T_\pi^\sigma Q
    \coloneqq (1 - \sigma) T_\pi Q + \sigma (T_\pi Q + w)
    = T_\pi Q + \sigma w
    .
\end{equation}
We can therefore write update (\ref{eq:qsigma_update}) in vector notation as
\begin{equation}
    \label{eq:qsigma_sequence}
    Q_{t+1} = (1-\alpha_t) Q_t + \alpha_t ( T_\pi Q_t + \sigma w_t)
    .
\end{equation}
\begin{theorem}
    \label{theorem:convergence}
    Assume $\sum_{t=0}^\infty \alpha_t = \infty$ and $\sum_{t=0}^\infty \alpha_t^2 < \infty$.
    For $\gamma < 1$, the sequence $Q_t$ defined by (\ref{eq:qsigma_sequence}) converges to $Q^\pi$ almost surely.
\end{theorem}
\proofglue
\begin{proof}
    Iteration~(\ref{eq:qsigma_sequence}) is already in a form where Proposition~4.4 of \citet{bertsekas1996neuro} is applicable;
    $T_\pi$ is a contraction mapping with respect to the maximum norm, and $\sigma w_t$ is zero-mean noise.
    Furthermore, ${\mathbb{E}[(\sigma w_t)^2] \leq E[w_t^2]}$ because $\sigma \in [0,1]$, and $E[w_t^2]$ is bounded by a constant since $\mathcal{S}$ and $\mathcal{A}$ are finite sets.
    Consequently, under the assumed stepsize conditions, the sequence $Q_t$ converges to $Q^\pi$ (the fixed point of $T_\pi$) almost surely.
\end{proof}
\proofglue
From our derived Q($\sigma$) operator (\ref{eq:qsigma_operator}), we see that $\sigma$ primarily functions as a noise attenuation parameter;
however, it does not affect the expected update, which is inherently related to the Bellman operator.
It follows that the contraction rate and fixed point of Q($\sigma$) are identical to those of the Bellman operator, and that these properties are independent of $\sigma$.
\subsection{Variance}

Theorem~\ref{theorem:convergence} suggests that small values of $\sigma$ have a beneficial variance-reduction effect, with no detrimental effects on convergence;
however, the abstract noise process $w_t$ makes it difficult to quantify this effect.
In this section, we derive the exact variance of Q($\sigma$) by extending existing results for Sarsa and Expected Sarsa from \citet{SeijenExpectedSARSA}.
\begin{theorem}
    \label{theorem:variance}
    Let $v_t$ and $\hat{v}_t$ be the variances of Expected Sarsa and Sarsa, respectively.
    The variance of Q($\sigma$) is given by the expression
    \begin{equation}
        \label{eq:qsigma_variance}
        \Var(v_t) + \sigma^2 (\Var(\hat{v}_t) - \Var(v_t))
        ,
    \end{equation}
    and this quantity is minimized when $\sigma = 0$.
\end{theorem}
\proofglue
\begin{proof}
    From the definition of Q($\sigma$) as a linear interpolation, we can deduce that its variance has the form
    \begin{equation}
        \label{eq:variance_formula}
        \sigma^2 \Var(\hat{v}_t) + (1 - \sigma)^2 \Var(v_t) + 2 \sigma ( 1 - \sigma) \Cov(\hat{v_t}, v_t)
        .
    \end{equation}
    The Sarsa and Expected Sarsa updates share the same expected value \citep{SeijenExpectedSARSA};
    let it be denoted by $\mu_t$.
    We show algebraically that $\Cov(\hat{v_t}, v_t) = \Var(v_t)$ by comparing the exact calculation $\mathbb{E}[(\hat{v_t} - \mu_t)(v_t - \mu_t)]$ to the expression for $\Var(v_t)$ derived by \citet{SeijenExpectedSARSA}.
    Therefore, expression~(\ref{eq:variance_formula}) reduces to
    \begin{equation*}
        \sigma^2 \Var(\hat{v_t}) + [(1 - \sigma)^2 + 2 \sigma ( 1 - \sigma)] \Var(v_t)
        = \Var(v_t) + \sigma^2 (\Var(\hat{v}_t) - \Var(v_t))
        ,
    \end{equation*}
    which is the desired result in expression~(\ref{eq:qsigma_variance}).
    From \citet{SeijenExpectedSARSA}, $\Var(\hat{v}_t) - \Var(v_t) \geq 0$, and hence expression~(\ref{eq:qsigma_variance}) is minimized when $\sigma=0$.
\end{proof}
\proofglue
Theorem~\ref{theorem:variance} shows that the variance of Q($\sigma$) is minimal when $\sigma=0$ and increases monotonically, which is interesting because it supports the view of $\sigma$ as a noise attenuation parameter from Theorem~\ref{theorem:convergence}.
Furthermore, since the expected update is unaffected by the choice of $\sigma$, it cannot represent a bias-variance trade-off;
the variance can be minimized with no apparent drawback.
Our theory indicates that there must be a different explanation for the empirical success of Q($\sigma$).

\tightsection{Adaptive Tree Backup (ATB) Algorithms}

Q($\sigma$) does not address the standard bias-variance trade-off for temporal-difference learning, since the choice of $\sigma=0$ minimizes variance without impacting bias.
Why, then, did \cite{de2018multi} observe empirical improvement when employing an intermediate $\sigma$-value?
We conjecture that Expected Sarsa ($\sigma=0$) struggles to overcome initialization error early in training (during which the majority of the state-action space is unexplored).
Until every action has been sampled at least once in a given state, Expected Sarsa mixes one or more nonsensical value estimates into its full backup, potentially resulting in severe inaccuracies.
In contrast, a sample method like Sarsa ($\sigma=1$) is likely to visit previously taken actions (simply by definition of a probabilistic policy), thereby bootstrapping sooner from more accurate estimates, but also with additional noise due to sampling.
This phenomenon helps explain why the time-decayed schedule for $\sigma$ proposed by \citet{de2018multi} is effective;
excluding incorrect value estimates ($\sigma=1$) is important when most state-action pairs have not been visited, but minimizing variance ($\sigma=0$) becomes more important once all estimates are relatively accurate.

If our hypothesis is correct, then methods that dynamically increase the effective backup width based on state-action pair visitations should generally perform better than Q($\sigma$) with a fixed $\sigma$-value or time-annealed schedule.
We call these Adaptive Tree Backup (ATB) methods, as they generalize Tree Backup \citep{precup2000eligibility} to arbitrary backup weightings.
For a given transition $(s_t,a_t,r_t,s_{t+1}, a_{t+1})$, ATB algorithms can be written in the generic form
\begin{equation}
    \label{eq:atb}
    Q(s_t,a_t) \gets~(1-\alpha_t) Q(s_t,a_t) + \alpha_t \left( r_t + \gamma \sum_{a' \in \mathcal{A}} c_t(s_{t+1}, a') Q(s_{t+1},a') \right)
    ,\quad
    \mathrm{subject~to}~\smash{\sum_{a \in \mathcal{A}}} c_t(s, a) = 1,
    \forall~s \in \mathcal{S},
\end{equation}
where each $c_t(s,a)$ is a nonnegative (possibly random) coefficient.
Note that if
$c_t(s,a) = (1-\sigma) \pi(a|s) + \sigma \mathbb{I}_{a=a_{t+1}}$,
then we recover Q($\sigma$) exactly.
However, the generality of the coefficients $c_t(s,a)$ permits myriad possibilities beyond the limited, one-dimensional spectrum spanned by Q($\sigma$).

The backup associated with update (\ref{eq:atb}) admits the correct fixed point $Q^\pi$ only if $\E[c_t(s,a)] = \pi(a|s)$.
Even so, if we ensure that $\E[c_t(s,a)] \to \pi(a|s)$ as $t \to \infty$, then the ATB method will eventually converge to $Q^\pi$ according to Theorem~\ref{theorem:convergence}.
We subsequently discuss two promising variants that do this.

\tightsubsection{Count-Based ATB}
\label{sect:count_atb}

To match the intuition that an effective ATB strategy should ignore unvisited state-action pairs, we develop a method that weights state-action pairs according to their relative frequency of appearance.
This can be accomplished by tracking a count $n(s,a)$ for each state-action pair $(s,a)$ and then normalizing it as a proportion.
This variant, which we call \emph{Count-Based} ATB, defines the backup coefficients as
\begin{equation}
    c_t(s,a) = \frac{n(s,a)}{\sum\limits_{a' \in \mathcal{A}} n(s,a')}
    .
\end{equation}
By the law of large numbers, $c_t(s,a) \to \pi(a | s)$ after infinitely many visitations to state $s$, and Theorem~\ref{theorem:convergence} with $\sigma=0$ applies at the limit.
Beyond convergence to $Q^\pi$, there are several reasons why this particular ATB formulation is appealing.
First, the count-based backup amounts to a maximum-likelihood $1$-step estimate of $Q(s,a)$, and in this sense is the ``most reasonable'' backup to perform based on previously observed experiences.
Furthermore, frequency counts implicitly emphasize value estimates that are more reliable, while excluding those that have not changed since initialization.
This means that Count-Based ATB automatically transitions from Sarsa to Expected Sarsa, with a different effective rate for each state, which could not be achieved easily with Q($\sigma$) and does not require a user-specified hyperparameter schedule.

Unfortunately, these benefits are not without some drawbacks.
Although the algorithm theoretically converges to $Q^\pi$ in the limit, it is extremely unlikely that $c(s,a) = \pi(a|s)$ after any finite amount of training.
This means that Count-Based ATB may tend towards a fixed point other than $Q^\pi$ in practice, which is undesirable even if it does represent a maximum-likelihood estimate according to past experience.
Another problem---which is much more significant---occurs when the agent switches its policy from $\pi$ to a different policy $\pi'$ during training.
At this point, all of the previously collected counts $n(s,a)$ will no longer reflect the correct on-policy distribution.
The agent would either need to reset the counts to zero (thus discarding useful information), or wait many timesteps for the counts to accurately reflect the new policy $\pi'$.
Both of these options are inefficient, motivating us to search for a better alternative in the next subsection.

\tightsubsection{Policy-Based ATB}
\label{sect:policy_atb}

We can resolve the issues of Count-Based ATB by directly utilizing knowledge of the policy $\pi$, as do Q($\sigma$), Expected Sarsa, and Tree Backup.
If some state-action pair $(s,a)$ is visited for the first time, then we know that the eventual value of $c_t(s,a)$ will be $\pi(a|s)$ according to Count-Based ATB.
We can greatly accelerate convergence by immediately assigning $c_t(s,a) = \pi(a|s)$ after this first visitation.
Let $u$ denote the unit step function such that $u(n) = 1$ if $n > 0$ and $u(n) = 0$ otherwise.
The backup coefficients for this new variant, which we call \emph{Policy-Based} ATB, become
\begin{equation}
    \label{eq:policy_atb}
    c_t(s,a) = \frac{u(n(s,a))}{\sum\limits_{a' \in \mathcal{A}} u(n(s,a'))} \pi(a|s)
    .
\end{equation}
This definition clearly solves the two problems of Count-Based ATB.
Unvisited state-action pairs are still ignored, as desired, but after each pair has been sampled at least once, the algorithm becomes equivalent to Expected Sarsa and starts converging to $Q^\pi$.
This effectively bypasses the infinite-visitation requirement of Count-Based ATB.
We can also see that the weighted backup instantly reflects changes to the policy, thanks to the explicit dependency on $\pi$ in definition~(\ref{eq:policy_atb}).
For these reasons, we expect Policy-Based ATB to learn significantly faster than Count-Based ATB in a tabular setting;
however, Policy-Based ATB may be harder to combine with function approximation in high-dimensional MDPs, where the need for generalization makes it difficult to determine whether a state should be considered visited or unvisited.
Additionally, the reliance on an explicit policy model $\pi(a|s)$ may be restrictive in some settings.
We therefore foresee useful roles for both ATB variants.

\section{Experiments and Conclusion}

\begin{wrapfigure}{r}{0.6\textwidth}
    \centering
    \hfill
    \includegraphics[width=0.29\textwidth]{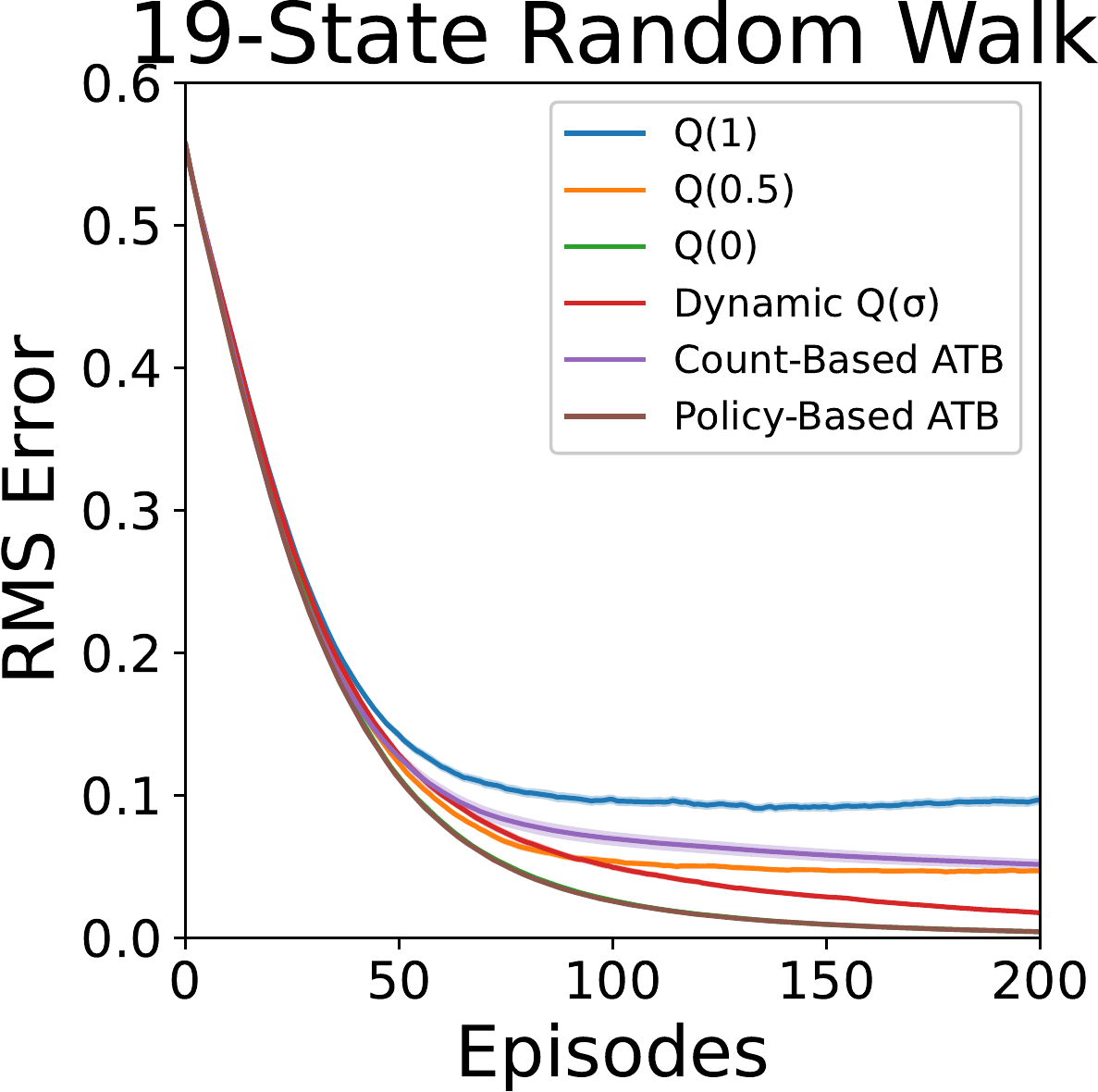}
    \hfill
    \includegraphics[width=0.29\textwidth]{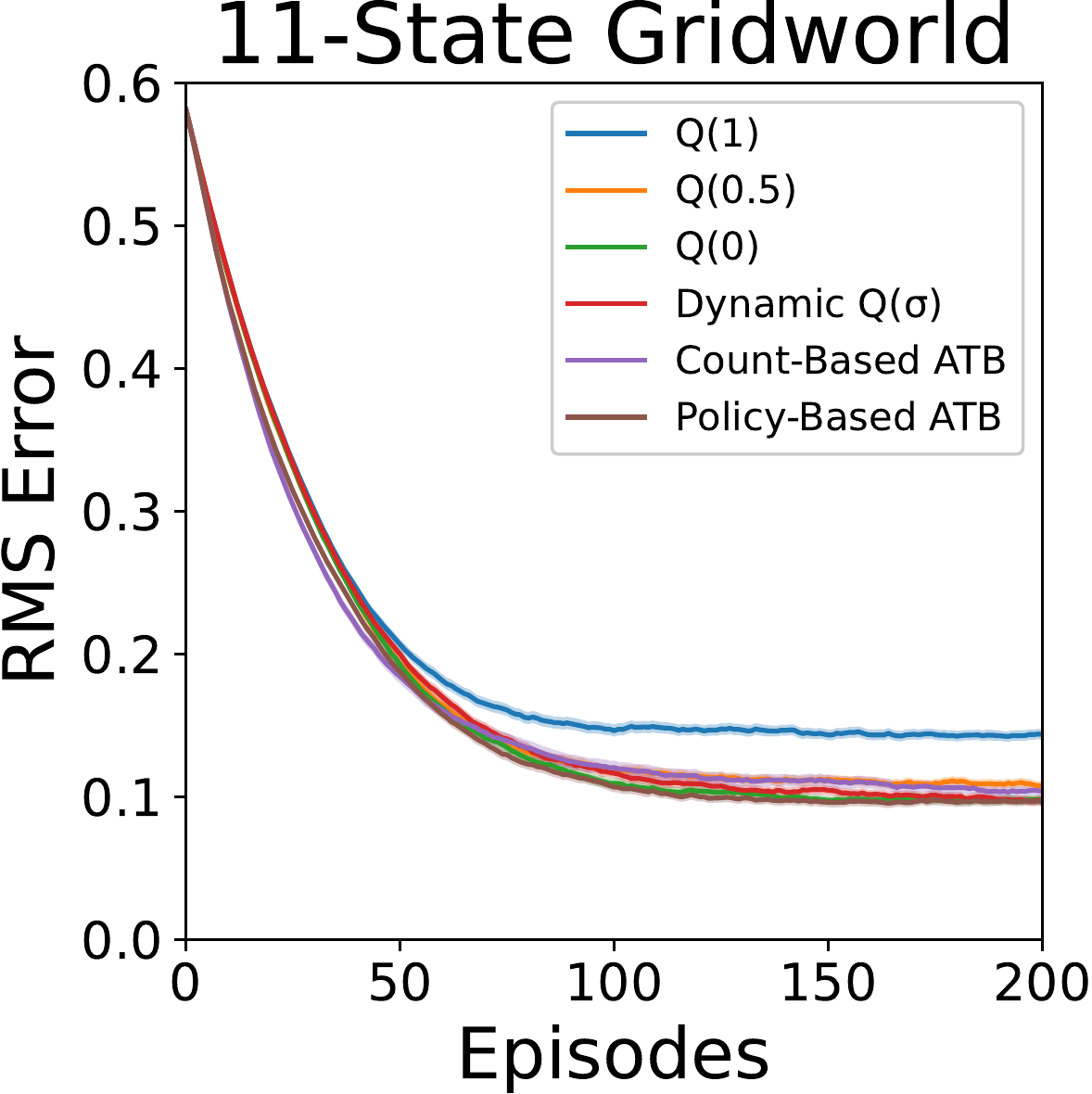}
    \hfill
    \caption{
        Comparison of our ATB methods against Q($\sigma$) with fixed $\sigma$-values and the dynamic schedule proposed by \citet{de2018multi}.
    }
    \label{fig:experiments}
\end{wrapfigure}

To test whether adaptive strategies can outperform fixed $\sigma$-values and the handcrafted schedule proposed by \citet{de2018multi}, we compare our two ATB variants against Q($\sigma$) in a 19-state deterministic random walk environment \citep{sutton2018reinforcement} and an 11-state stochastic gridworld environment \citep{russel2010}.
For all methods, we fix $\alpha_t = 0.4$ and $\gamma = 1$.
We train the agents for 200 episodes and plot the root-mean-square (RMS) error of the estimated value function $Q$ relative to $Q^\pi$ (Figure~\ref{fig:experiments}).
We plot the error after each episode, averaging over 50 independent trials, where the shaded regions represent 99\% confidence intervals.
Policy-Based ATB performs strongly, learning significantly faster than all of the methods except Q($0$), and even surpassing the handcrafted, dynamic $\sigma$ schedule of \citet{de2018multi}---an impressive feat for an automatic method.
Count-Based ATB outperforms Q($1$), but its asymptotic performance in the deterministic random walk deteriorates due to the fixed-point bias we discussed earlier;
however, this seems to have a less detrimental effect in the stochastic gridworld.
Surprisingly, in contrast to the results of \citet{de2018multi}, we find that $\sigma=0$ is the best choice of fixed $\sigma$-value in both environments.

The empirical success of our adaptive methods---particularly Policy-Based ATB---has important implications.
First, it shows that data-driven adaptive strategies for adjusting the backup width can be more effective than Q($\sigma$) with a fixed $\sigma$-value or a pre-specified schedule for annealing $\sigma$.
In addition, since both of our methods gradually expand the effective backup width as new state-action pairs are visited, they further strengthen our hypothesis that initialization error in the value estimates is an important consideration when determining backup width.
This also helps formalize the intuition of \citet{de2018multi} that backup width should increase over time.
Finally, these results show that $\sigma=0$ is a good choice in practice, corroborating our theoretical predictions that it minimizes variance without negatively impacting the bias or contraction rate of Q($\sigma$).
Although $\sigma$ does not represent a bias-variance trade-off in temporal-difference learning, it is clear that $\sigma$ does play an important role in learning, and that analyzing methods in terms of how well their backups can balance initialization error with variance is a promising pathway to further algorithmic improvements.

\bibliographystyle{plainnat}
\small
\bibliography{references}

\end{document}